\documentclass{article}
\usepackage{ijcai18}

\usepackage{times}
\usepackage{soul}
\usepackage{url}
\usepackage[hidelinks]{hyperref}
\usepackage[utf8]{inputenc}
\usepackage[small]{caption}
\usepackage{graphicx}

\title{Addition of Code Mixed Features to Enhance the Sentiment Prediction of Song Lyrics \thanks{This work was presented at 1st Workshop on Humanizing AI (HAI) at IJCAI'18 in Stockholm, Sweden.}}

\author{
Gangula Rama Rohit Reddy$^1$,
Radhika Mamidi$^2$
\\
$^1$,$^2$ Language Technologies Research Center\\International Institute of Information Technology, Hyderabad (India) \\
ramarohitreddy.g@research.iiit.ac.in,
radhika.mamidi@iiit.ac.in
}

\begin{document}

\maketitle

\begin{abstract}
  Sentiment analysis, also called opinion mining, is the field of study that analyzes people’s opinions, sentiments, attitudes and emotions. Songs are important to sentiment analysis since the songs and mood are mutually dependent on each other. Based on the selected song it becomes easy to find the mood of the listener, in future it can be used for recommendation. The song lyric is a rich source of datasets containing words that are helpful in analysis and classification of sentiments generated from it. Now a days we observe a lot of inter-sentential and intra-sentential code-mixing in songs which has a varying impact on audience. To study this impact we created a Telugu songs dataset which contained both Telugu-English code-mixed and pure Telugu songs. In this paper, we classify the songs based on its arousal as exciting or non-exciting. We develop a language identification tool and introduce code-mixing features obtained from it as additional features. Our system with these additional features attains 4-5\% accuracy greater than traditional approaches on our dataset. 
\end{abstract}

\section{Introduction}

Code Mixing is a natural phenomenon of embedding linguistic units such as phrases, words or morphemes of one language into an utterance of another. Code-mixing is widely observed in multilingual societies like India. Telugu is an agglutinative Dravidian language spoken widely in India. It is the third most popular language in India after Hindi and Bengali\footnote{\url{https://www.ethnologue.com/statistics/size}}. The rapid increase of Telugu movie songs content on the Internet opened up tremendous potential for research in sentiment and mood analysis. Listening to songs has a strong relation with the mood of the listener. Song lyrics can be analyzed as a dataset consisting of thoughts, moods, emotions, believes or feelings related to certain topics. Lyrics provides a much broad view of song than simpler forms of meta data like title, artist, year and hence contains true content. Words or Lyrics is the soul of song and songs with powerful lyrics cause arousal or excitement in the people. A particular mood can drive us to select some song and a song can invoke some sentiment, valence or arousal in us, which can change our mood. Generally songs are classified as positive or negative which do not reflect the excitement or arousal in the mood of the listener. In this paper, we created a dataset which is annotated as exciting or non-exciting based on the arousal in its lyrics and classify the songs. In the present day, we find a lot of code-mixing in song lyrics and it has significant effect on the mood driven by that song. In this paper, we showed that including the code-mixed features like number of words from other language present in the song, average word length, number of sentences present etc. as additional features to the classification algorithms attains 4-5\% high accuracy. In order to obtain these code-mixed features, a language identification tool is used.

\section{Related Work}
In English, sentiment analysis systems have been applied to many different kinds of texts
including customer reviews \cite{DBLP:conf/aaai/HuL04,DBLP:books/daglib/0036864,DBLP:conf/aaai/2004}, newspaper headlines \cite{article1} etc. Often these systems have to cater to the specific needs of the text such as formality versus informality, length of utterances, etc.

Sentiment analysis of Telugu texts has several challenges. It is morphologically complex language. Very little work is done on sentiment analysis in Telugu. Sentiment analysis systems have been applied to different kinds of Telugu texts including Song Lyrics \cite{DBLP:conf/ijcai/AbburiAGM16}, News \cite{DBLP:conf/ijcai/MukkuCM16} and Reviews\cite{GANGULA18.146}.
Very little work is done on code mixing in Telugu. Part-of-Speech tagging for code mixed English-Telugu social media data was done by \cite{DBLP:conf/cicling/NelakuditiJM16}. in which authors use manually extracted features for tagging the data.

\section{Dataset}
In order to gather a dataset of unique song lyrics, we mined song lyrics from two websites viz. lyricsing.com and telugulyrics.co.in. After mining the lyrics, we cleaned them of html tags and other extraneous text. This created a dataset of 1744 Telugu song lyrics.
Several papers\cite{DBLP:conf/ismir/HanRDH09,DBLP:conf/ismir/HuD10,DBLP:conf/ismir/HuCY09} make use of Russell's model, a two-dimensional model to rate emotions, most of which focus on valence than arousal. This approach relies on assigning emotions based on their locations on the two-dimensional co-ordinate system. This model defines two dimensions, viz. Valence and Arousal. Valence is a subjective feeling of pleasantness or unpleasantness. Arousal is a subjective state of feeling activated or deactivated. Each song in our dataset is annotated based on its arousal as exciting or non-exciting by the annotators. Most of the songs in Telugu movies tend to be positive, so when creating a dataset there would be very few negative songs present in it causing imbalance. Due to the above reason, we decided not to consider valence.    

\begin{figure}[h!]
  \centering
  \includegraphics[width=0.5\textwidth]{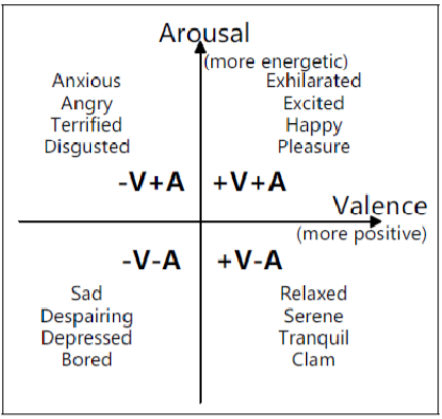}
  \caption{Rusell's Model for Mood}
\end{figure}

For each song in our dataset, firstly annotators went through the lyrics and annotated them. But annotating only based on lyrics would be misleading as it depends on the situational context in the movie. So, annotators learned the situation of the song in the movie and corrected the wrong annotations. Finally after multiple meetings and discussions with them, a kappa score of 0.86 is achieved.
Table 1 shows the statistics of our dataset.
\begin{table}[h!]
\centering
\label{my-label}
\begin{tabular}{|l|l|l|}
\hline
Total songs & Exciting & Non-Exciting \\ \hline
1744        & 830     & 914         \\ \hline
\end{tabular}
\caption{Corpus Statistics}
\end{table}

The dataset is available at \url{https://github.com/Rama-007/song_lyrics-dataset} .
\section{Code Mixed Feature Extraction}
Language identification is the process of dividing words into one of the mentioned languages. It is important for the extraction of code mixed features. We use the following method for language identification.
\subsection{Language Identification}
For language identification, we used the method mentioned in \cite{DBLP:conf/cicling/NelakuditiJM16}. A CRF model is implemented for language identification with the following features.
\begin{itemize}
\item lexical feature:
\begin{itemize}
\item word
\end{itemize}
\item sub-lexical features:
\begin{itemize}
\item Prefix, suffix character strings
\item Infix character strings
\item Presence of Post positions
\item Prefix, Suffix character strings of neighboring words
\end{itemize}
\item other features:
\begin{itemize}
\item length of the word
\item neighboring words
\item presence in English dictionary 
\end{itemize}
\end{itemize}
This model achieved an accuracy of 81.4\% in language identification.

Once language of each word is identified using this model, we extract the following features and use them as our code-mixed features.
\begin{itemize}
\item Number of non-Telugu words present in the song.
\item Average word length of non-Telugu words present in the song.
\item Number of non-Telugu sentences present in the song.
\item Average length of non-Telugu sentences present in the song.
\end{itemize}

\section{CMNN Model}
In this section, we introduce the overall architecture of our code mixed neural network (CMNN) model. Figure 2 depicts the proposed architecture of CMNN model.
\begin{itemize}
\item \textbf{Embedding Layer}

Our model accepts song lyrics and its label as a training instance. Each song lyric is represented as a fixed length sequence in which we pad all sequences to the maximum length. Let L be maximum length. Subsequently each sequence is converted into a sequence of low-dimensional vectors via the embedding layer. Word2Vec embeddings are used for this purpose.

\item \textbf{Convolution Layer}

The sequence of word embeddings obtained from the embedding layer is passed into a 1D convolution layer. This layer is beneficial because it helps to extract local features from the sequence.

\item \textbf{Long Short-Term Memory (LSTM)}

The output of convolution layer is passed into a long short-term memory. LSTM outputs a hidden vector $h_{t}$ that reflects the semantic representations at position t. To select the final representation of the song lyric, a temporal mean pool is applied to all LSTM outputs. 

\item \textbf{Code mixed features}

The code mixed features are extracted for each song as mentioned in Section 4 and they are concatenated with final representation of LSTM layer and is passed into fully-connected hidden layer.

\item \textbf{Fully-connected Hidden Layer}

The concatenated vector is passed through a fully connected hidden layer defined as follows:

\begin{center}
$h_{out}$ = f($W_{h}$([e, $s_{1}$, $s_{2}$, $s_{3}$, $s_{4}$]))+$b_{h}$
\end{center} 

where f(.) is a non-linear activation such as tanh or relu, $W_{h}$ and $b_{h}$ are the parameters of the hidden layer. e is the final representation obtained from temporal mean pooling and $s_{1}$, $s_{2}$, $s_{3}$, $s_{4}$ are the code mixed features.

Finally, we pass $h_{out}$ into a softmax layer and predict song's class.

\end{itemize}

\begin{figure}[h!]
  \centering
  \includegraphics[width=0.5\textwidth]{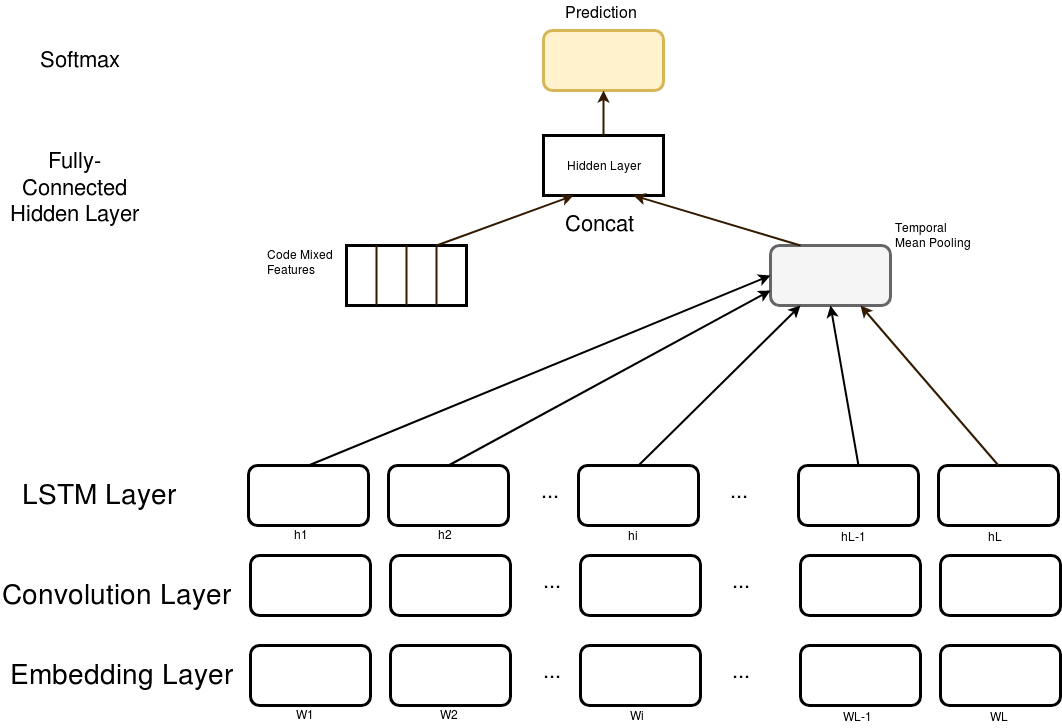}
  \caption{CMNN Model Architecture}
\end{figure}

\section{Experimental Evaluation}
In this section, we describe our experimental procedure and results.
\subsection{Experimental Setup}
We use 5-fold cross validation to evaluate all systems with 64/16/20 split for train, development and test sets. We train all the models for 10 epochs and select the best model based on the performance on the development set.
\subsection{Evaluation Metric}
The evaluation metric used is accuracy score.
\subsection{Baselines and Implementation Details}
In this section, we discuss the algorithms that are used as baselines for our model.
\begin{itemize}
\item \textbf{Naive Bayes} - We report the results of Bayesian classifier with tf-idf vector of songs as input.
\item \textbf{SVM} - We report the results of Support Vector Machine classifier with tf-idf vector of songs as input.
\item \textbf{CNN} - We implemented a CNN model using 1D convolutions. We use a filter width of 3. The outputs from CNN model are passed through a fully connected hidden layer and finally through softmax layer. 
\item \textbf{LSTM} - We implemented LSTM model where the outputs from LSTM model are passed through mean pooling layer and then through fully connect hidden layer and finally through softmax layer.
\item \textbf{CMNN Model}
\begin{itemize}
\item Without Code-mixed Features
\item With Code-mixed Features
\end{itemize}

\end{itemize}

\subsection{Results and Analysis}
Table 2 reports the results of all the models. We can observe that our model with code-mixed features outperforms the other models by almost 4-5\% in accuracy. This is because, in present day Telugu songs the arousal in song increases with the amount of code-mixing in it. So providing the model with this additional information helps the model in prediction.
\begin{table}[h!]
\centering
\label{my-label}
\begin{tabular}{|l|l|l|}
\hline
ID & Model                                                                              & Accuracy \\ \hline
1  & Naive Bayes                                                                        & 53.57\%     \\ \hline
2  & Naive Bayes with code-mixed features                                                                       & 62\%     \\ \hline
3  & SVM                                                                                & 60\%     \\ \hline
4  & SVM with code-mixed features                                                                               & 69\%     \\ \hline
5  & CNN                                                                                & 67\%     \\ \hline
6  & CNN with code-mixed features                                                                               & 66.2\%     \\ \hline
7  & LSTM                                                                               & 68\%   \\ \hline
8  & LSTM with code-mixed features                                                                               & 69.3\%   \\ \hline
9  & \begin{tabular}[c]{@{}l@{}}CMNN model \\ without code\\ mixed features\end{tabular} & 71.2\%     \\ \hline
10  & \begin{tabular}[c]{@{}l@{}}CMNN model \\ with code mixed features\end{tabular}      & 76.6\%   \\ \hline
\end{tabular}
\caption{Experimental results of all compared models.}
\end{table}

\section{Conclusion}
In this paper, we presented the method of creation and annotation of Telugu song lyrics dataset annotated based on its arousal. We proposed the method of extraction of code-mixed features from song lyrics using language identification tool and incorporating them in our model to achieve a significant 4-5\% increase in accuracy.

\bibliographystyle{named}
\bibliography{ijcai18}

\end{document}